\documentclass[11pt,a4paper]{article}
\usepackage[hyperref]{emnlp2020}
\usepackage{times}
\usepackage{latexsym}

\def\urltilda{\kern -.0em\lower .7ex\hbox{\~{}}\kern .04em}
\usepackage{microtype}

\aclfinalcopy % Uncomment this line for the final submission
\def\aclpaperid{894} %  Enter the acl Paper ID here

\usepackage{tabularx}
\usepackage{wasysym}
\usepackage{graphicx}
\usepackage{multirow}

\title{Can Automatic Post-Editing Improve NMT? }

\author{ Shamil Chollampatt \\ Rakuten, Inc. \\ shamil.chollampatt@rakuten.com \And Raymond Hendy Susanto \\ Rakuten, Inc. \\ raymhs91@gmail.com \AND  Liling Tan  \\ Rakuten, Inc.  \\ liling.tan@rakuten.com \And Ewa Szymanska \\ Rakuten, Inc. \\ ewa.szymanska@rakuten.com}

\date{}

\begin{document}
\maketitle
\begin{abstract}

Automatic post-editing (APE) aims to improve machine translations, thereby reducing human post-editing effort. APE has had notable success when used with statistical machine translation (SMT) systems but has not been as successful over neural machine translation (NMT) systems. This has raised questions on the relevance of APE task in the current scenario. However, the training of APE models has been heavily reliant on large-scale artificial corpora combined with only limited human post-edited data. We hypothesize that APE models have been
underperforming in improving NMT translations due to the lack of adequate supervision. To ascertain our hypothesis, we compile a larger corpus of human post-edits of English to German NMT. We empirically show that a state-of-art neural APE model trained on this corpus can significantly improve a strong in-domain NMT system, challenging the current understanding in the field. We further investigate the effects of varying training data sizes, using artificial training data, and domain specificity for the APE task. We release this new corpus under CC BY-NC-SA 4.0 license at \url{https://github.com/shamilcm/pedra}.

\end{abstract}

%%%%%%%%%%%%%%%%%%%%%%%%
\section{Introduction}
%%%%%%%%%%%%%%%%%%%%%%%%

\begin{figure}
\centering
\fbox{\includegraphics[width=0.45\textwidth]{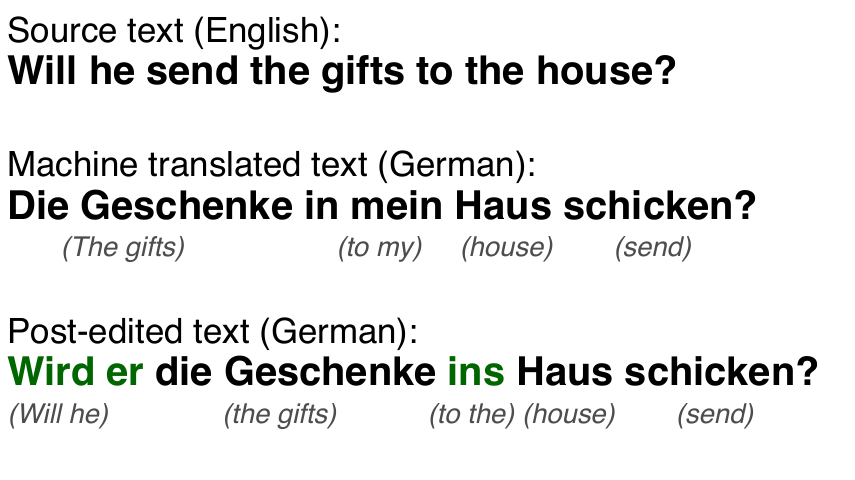}}
\caption{An example of post-editing given the source text in English and the translated text in German. }
\label{fig:post-editing}
\end{figure}

Automatic Post-Editing (APE) aims to reduce manual post-editing effort by automatically fixing errors in the machine-translated output. \citet{knight1994automated} first proposed APE to cope with systematic errors in selecting appropriate articles for Japanese to English translation. Earlier application of statistical phrase-based models for APE treated it as a monolingual re-writing task without considering the source sentence \cite{simard2007statistical,bechara2011statistical}. Modern APE models take the source text and machine-translated text as input and output the post-edited text in the target language (see Figure \ref{fig:post-editing}).

APE models are usually trained and evaluated in a \textit{black-box} scenario where the underlying MT model and the decoding process are inaccessible making it difficult to improve the MT system directly. APE can be effective in this case to improve the MT output or to adapt its style or domain.

Recent advancement of APE has shown remarkable success on statistical machine translation (SMT) outputs \cite{junczysdowmunt2018msuedin,correia2019simple} even when trained with limited number of post-edited training instances (generally ``triplets'' consisting of \textit{source}, \textit{translated}, and \textit{post-edited} segments), with or without additional large-scale artificial data \cite{junczysdowmunt2016loglinear,negri2018escape}. Substantial improvements have been reported especially on English-German (EN-DE) WMT APE shared tasks on SMT \cite{bojar2017findings,chatterjee2018findings}, when models were trained with fewer than 25,000 human post-edited triplets. However, on NMT, strong APE models have failed to show any notable improvement \cite{chatterjee2018findings,chatterjee2019findings,ive2020post} when trained on similar-sized human post-edited data. This has led to questions regarding the usefulness of APE with current NMT systems that produce improved translations compared to SMT. \citet{junczysdowmunt2018msuedin} concluded that the results of the WMT'18 APE (NMT) task ``might constitute the end of neural automatic post-editing for strong neural in-domain systems" and that ``neural-on-neural APE might not actually be useful''. Contrary to this belief, we hypothesize that a competitive neural APE model still has potential to further improve strong state-of-the-art in-domain NMT systems when trained on adequate human post-edited data.

We compile a new large post-edited corpus, \textit{SubEdits}, which consists of actual human post-edits of translations of drama and movie subtitles produced by a strong in-domain proprietary NMT system. We use this corpus to train a state-of-the-art neural APE model \cite{correia2019simple}, with the goal of answering the following three research questions to better assess the relevance of APE going forward:
\begin{itemize}
    \item Can APE substantially improve in-domain NMT with adequate data size?
    \item How much does artificial APE data help?
    \item How significant is domain shift for APE? 
\end{itemize}

\noindent \textbf{Spoilers:} Through automatic and human evaluation, we confirm our hypothesis that, in order to notably improve over the original NMT output (``do-nothing'' baseline), state-of-the-art APE models need to be trained on a larger number of human post-edits, unlike the case with SMT. Training on datasets of sizes in the scale of those from the WMT APE tasks, even with large-scale in-domain artificial APE corpora, leads to underperformance. Our experimental results also highlight that APE models are highly sensitive to domain differences. To effectively exploit out-of-domain post-edited corpora such as SubEdits in other domains, it has to be carefully mixed with available in-domain data.

%%%%%%%%%%%%%%%%%%%%%%%%%%%
\section{SubEdits Corpus}
%%%%%%%%%%%%%%%%%%%%%%%%%%%

\begin{table}[t]
    \centering
    \small
    \begin{tabular}{|l|c|r|c|}
    \hline
    & \textbf{Lang.} & \textbf{Size} & \textbf{Domain}
    \\ 
    \hline
    \multicolumn{4}{|l|}{\textbf{Human post-edited corpora}} \\
    \hline
    & & & \\[-1em]
    QT21 & \multirow{2}{*}{EN-LV} & \multirow{2}{*}{21K} & Life\\
    \cite{specia2017translation}  & & & Sciences \\
     & & & \\[-1em]
    \hline 
     & & & \\[-1em]
    WMT'18 \& '19 APE & \multirow{2}{*}{EN-DE} & \multirow{2}{*}{15K} & \multirow{2}{*}{IT} \\
    \cite{chatterjee2018findings}  & & & \\
    & & &  \\[-1em]
    \hline
    & & &  \\[-1em]
    WMT'19 APE  & \multirow{2}{*}{EN-RU} & \multirow{2}{*}{17K} & \multirow{2}{*}{IT} \\
    \cite{chatterjee2019findings}  & & & \\
    & & &  \\[-1em]
    \hline
     & & & \\[-1em]
    APE-QUEST  &   EN-NL & 11K & \multirow{3}{*}{Legal}\\
    \cite{ive2020post} & EN-FR & 10K &   \\
                             & EN-PT & 10K &  \\    
    \hline
     & & & \\[-0.7em]
    SubEdits  (\textit{this work}) & EN-DE & 161K & Subtitles \\
    \hline
     & & & \\[-1em]

    \multicolumn{4}{|l|}{\textbf{Artificial corpora}} \\
    \hline
     & & & \\[-1em]
    eSCAPE  &   EN-DE & 7.2M & \multirow{3}{*}{Mixed}\\
    \cite{negri2018escape} & EN-IT & 3.3M &   \\
                             & EN-RU & 7.7M &  \\
    \hline
    \end{tabular} 
    \caption{APE corpora on NMT outputs and their sizes in terms of number of post-edited triplets. }
    \label{tbl:ape-data}
\end{table}

 Human post-edited corpora of NMT outputs from previous WMT APE shared tasks usually consist of fewer than 25,000 instances. Large-scale artificial corpora such as eSCAPE \cite{negri2018escape}, do not adequately cater to the primary APE objective of correcting systematic errors of the MT outputs since the pseudo ``post-edits'' are independent human-translated references often differing greatly from the MT output. Table \ref{tbl:ape-data} lists the real and artificial APE corpora on NMT outputs. 
Due to the paucity of larger human post-edited corpora on NMT outputs, a study of APE performance under sufficient supervised training data conditions was not possible previously. To enable such a study, we introduce the SubEdits EN-DE post-editing corpus with over 161K triplets of source sentences, NMT translations, and human post-edits of NMT translations.

\subsection{Corpus Collection}
SubEdits corpus is collected from a database of subtitles of a popular video streaming platform, Rakuten Viki (\href{https://www.viki.com/}{https://www.viki.com/}) Every subtitle segment had been originally manually transcribed and translated to English before translating it to German using a proprietary NMT system employed by the platform and specialized at translating subtitles. Viki community\footnote{\href{https://contribute.viki.com/}{https://contribute.viki.com/}} members who volunteer as subtitle translators would then post-edit the machine-translated subtitles to further improve it, if necessary.

\subsection{Corpus Filtering}
\label{subsec:data-filtering}
We use an adaptation of Gale-Church filtering \cite{tan2014manawi} used for machine translation for filtering the triplets. The global character mean ratio $r_c$ is computed as the ratio between the number of characters in the source and machine translated portions of the entire corpus. We remove triplets (\textit{src}, \textit{mt}, \textit{pe}) from the corpus where the ratio between the number of characters of source (\textit{src}) and post-edit (\textit{pe}) does not lie within a threshold range of $(1-t)r_c$ and $(1+t)r_c$ with $t=0.2$. We normalize punctuation\footnote{Using Moses \href{https://github.com/moses-smt/mosesdecoder/blob/master/scripts/tokenizer/normalize-punctuation.perl}{normalize-punctuation.perl} script.} and remove duplicate triplets. Among the triplets that share the same \textit{src} and \textit{mt} segments, we choose only the one with the longest \textit{pe}. Finally, we remove triplets that are not correctly identified with the respective source and target language using a language identification tool\footnote{\href{https://github.com/saffsd/langid.py}{https://github.com/saffsd/langid.py}} \cite{lui2012langid}. We set aside 10,000 triplets as development set and 10,000 triplets as test set. The final statistics are shown in Table \ref{tbl:dataset-stat}.

\begin{table}
\small
\begin{tabular}{|c|r|rrr|}
\hline
 & \multicolumn{1}{c|}{\textbf{No. of}}  & \multicolumn{3}{c|}{\textbf{ No. of tokens} } \\
 & \multicolumn{1}{c|}{\textbf{triplets}}  & \multicolumn{1}{c}{\textbf{src}} & \multicolumn{1}{c}{\textbf{mt}} & \multicolumn{1}{c|}{\textbf{pe}} \\
 \hline
Train & 141,413 & 1,432,247 & 1,395,211 & 1,423,257 \\
Dev & 10,000 & 101,330 & 98,581 & 100,795 \\
Test & 10,000 & 101,709 & 99,032 & 101,112 \\
\hline
\end{tabular}
\caption{Statistics of the SubEdits corpus}
\label{tbl:dataset-stat}
\end{table}

\section{BERT Encoder-Decoder APE Model}

BERT Encoder-Decoder APE \cite{correia2019simple} is a state-of-the-art neural APE model based on a Transformer model \cite{vaswani2017attentiom} with the encoder and decoder initialized with pre-trained multilingual BERT \cite{devlin2019bert} weights and fine-tuned on post-editing data.

A single encoder is used to encode both the source text and the machine-translated text by concatenating them with the separator token \texttt{[SEP]}. The encoder component of the model is identical to the original Transformer encoder initialized with pre-trained weights from the multilingual BERT.  For the decoder, \citet{correia2019simple} initialized the context attention weights with the corresponding BERT self-attention weights. Also, the weights of the self-attention layers of the encoder and decoder are tied. All other weights are initialized with corresponding weights from the same multilingual BERT model as well.

BERT Encoder-Decoder APE was shown to outperform other state-of-the-art APE models \cite{tebbifakhr2018multi,junczysdowmunt2018msuedin} on SMT outputs even in the absence of additional large-scale artificial data that competing models have used. An improved variant of this model with additional in-domain artificial data, despite being the winning submission of the recent WMT'19 APE EN-DE (NMT) task \cite{lopes2019unbabel}, only performed marginally better than the baseline NMT output. For the purpose of this study, we base our experiments on the BERT Encoder-Decoder APE architecture \citep{correia2019simple}.

\section{Experimental Setup}

\subsection{Model Hyperparameters}
For the BERT Encoder-Decoder model (\textbf{BERT Enc-Dec}), we use the implementation\footnote{\href{https://github.com/deep-spin/OpenNMT-APE}{https://github.com/deep-spin/OpenNMT-APE}} and model hyperparameters used by \citet{correia2019simple} and initialize the encoder and decoder with cased multilingual BERT (base) from Transformers\footnote{\href{https://github.com/huggingface/transformers}{https://github.com/huggingface/transformers}} library \cite{wolf2019huggingface}. The encoder and decoder follow the architecture of BERT (base) with 12 layers and 12 attention heads, an embedding size of 768, and a feed-forward layer size of 3072. We set the effective batch size to 4096 tokens for parameter updates. We train BERT Enc-Dec on a single NVIDIA Quadro RTX6000 GPU. Training on our SubEdits corpus took approximately 5 hours to converge. We validate and save checkpoints at every 2000 steps and use early-stopping (patience of 4 checkpoints) to select the model based on best perplexity. We use a decoding beam size of 5.

As a control measure, we compare BERT Enc-Dec against two vanilla Transformer APE models using automatic metrics. The Transformer APE models use BERT vocabularies and tokenization, and employ a single encoder to encode the concatenation $src$ and $mt$, but they are not initialized with pre-trained weights. The following are the descriptions of the two Transformer APE baselines:
\paragraph{TF (base)} A Transformer (base) \cite{vaswani2017attentiom} model with 6 hidden layers implemented in OpenNMT-py.\footnote{\href{https://github.com/OpenNMT/OpenNMT-py}{https://github.com/OpenNMT/OpenNMT-py}} The embedding size is 512 with 2048 feed-forward units. We use default learning parameters in OpenNMT-py: Adam optimizer with a learning rate of 2 and Noam scheduler. 
\paragraph{TF (BERT size.)} A bigger Transformer with the same number of layers, attention heads, embedding dimensions, hidden, and feed-forward dimensions as BERT Enc-Dec, but without any pre-training and tying of self-attention layers. All learning hyperparameters follow that of TF (base) model.

\subsection{Pre-processing and Post-processing}
\label{sec:prepost} 
SubEdits corpus contains HTML tags such as line breaks (\texttt{<br>}) and italic tags (\texttt{<i>}), and symbols denoting musical notes (\twonotes, \eighthnote) and segments often begin with hyphens (-). We applied several processing steps to make the data as close as possible to natural sentences on which BERT has been pre-trained on. The triplets with multi-line $src$, $mt$, and $pe$ containing \texttt{<br>} tags are split into separate training instances\footnote{We only separate at \texttt{<br>} when the \textit{src},\textit{mt}, and \textit{pe} contains same number of \texttt{<br>}  symbols.} and we remove italics and other HTML tags, musical note symbols, and leading hyphens. Thereafter, the input is tokenized with the BERT tokenization and word-piece segmentation in the Transformers library. During test time, we keep track of the changes made to input such as deletion of leading hyphens, music symbols, and italics tags, and splitting at \texttt{<br>} tags. After decoding, the outputs are detokenized and post-processed to re-introduce the tracked changes and evaluated.

\subsection{Evaluation}

We evaluate the models using three different automatic metrics: BLEU \cite{papineni2002bleu}, ChrF \cite{popovic2015ChrF}, and TER \cite{snover2006ter}. 
For our evaluation on SubEdits test set, differing from WMT APE task evaluation, we post-process and detokenize the outputs and use SacreBLEU\footnote{\href{https://github.com/mjpost/sacreBLEU}{https://github.com/mjpost/sacreBLEU}} \cite{post2018call} to evaluate BLEU and ChrF, and TERCOM\footnote{\href{http://www.cs.umd.edu/\urltilda snover/tercom/}{http://www.cs.umd.edu/\urltilda snover/tercom/}} to compute TER with normalization. Significance test is done by bootstrap re-sampling on BLEU with 1000 samples \cite{koehn2004statistical}. Additionally, we conduct human evaluation to ascertain the improvement of the BERT Enc-Dec APE model and to determine the human upper-bound performance for the SubEdits benchmark (see Section \ref{sec:humaneval}). 

We also compare the APE model on the canonical WMT APE dataset (Section \ref{sec:wmt-eval} and Table \ref{tbl:wmt}). We follow their evaluation method and use the released tokenized post-edited reference to compute BLEU, ChrF, and TER on the tokenized output.

\section{Results and Discussion}

\subsection{Proprietary In-domain NMT}
\label{sec:baseline-perf}

\begin{table}[t]
\centering
\small
%\newcolumntype{R}{>{\raggedleft\arraybackslash}X} 
%\begin{tabularx}{0.46\textwidth}{|p{2.3cm}|RRR|}
\begin{tabular}{|l|rrr|}
\hline
& \textbf{BLEU}$\uparrow$ & \textbf{ChrF}$\uparrow$ & \textbf{TER}$\downarrow$  \\
\hline
Proprietary NMT &  46.83 & 63.81 & 37.20\\
\hline
Google Translate            & 40.96 & 59.20 & 41.91 \\
Microsoft Translator        & 38.78 & 57.68 & 43.72 \\
SYSTRAN                     & 38.06 & 56.74 & 44.37  \\
%MT4             & 22.99 & 44.77 & 58.38 \\
\hline
\end{tabular}
\caption{Comparison of the proprietary NMT to leading commercial MT systems on an in-domain test set.}
\label{tbl:commercial}
\end{table}

\begin{table*}[t]
\centering
\small
\newcolumntype{R}{>{\raggedleft\arraybackslash}X}
\begin{tabularx}{0.97\textwidth}{|p{3.1cm}|R|R|R|R|R|R|R|}
\hline
    & \multicolumn{1}{c|}{\textbf{No. of}} &  \multicolumn{3}{c|}{ \textbf{Dev} } & \multicolumn{3}{c|}{ \textbf{Test} } \\
     \cline{3-8}
& \multicolumn{1}{c|}{\textbf{Params}} & \textbf{BLEU}$\uparrow$ &% $\Delta$ &
\textbf{ChrF}$\uparrow$ &%  $\Delta$ &
\textbf{TER}$\downarrow$ &%  $\Delta$ &
\textbf{BLEU}$\uparrow$ &%  $\Delta$ &
\textbf{ChrF}$\uparrow$ &%  $\Delta$ &
\textbf{TER}$\downarrow$  %$\Delta$ \\
\\
\hline
\textit{do-nothing} NMT  &  & 62.07 & 71.66 & 27.68 & 61.88 & 71.33 & 28.06 \\
~w/ TF (Base) APE    & 105.5M & 62.47 & 72.26 & 25.65 & 62.26 & 71.97 & 25.94 \\
~w/ TF (\textsc{bert} size.) APE & 290.4M & 62.04 & 72.04 & 25.73 & 61.62 & 71.65 & 26.14 \\
~w/ BERT Enc-Dec APE          & 262.4M & \textbf{64.88} & \textbf{74.94} & \textbf{23.29} & \textbf{64.53} & \textbf{74.71} & \textbf{23.72} \\
\hline
\end{tabularx}
\caption{Performance of APE models on the SubEdits test set. }
\label{tbl:baseline}
\end{table*}

We first assess the quality of an proprietary in-domain NMT system that is used for compiling the SubEdits corpus. We use it as a black-box system and use the evaluation results from Table \ref{tbl:commercial} to demonstrate that it is a strong baseline for studying APE performance on NMT outputs.

We compare the proprietary NMT system to three leading commercial EN-DE NMT systems: Google Translate, Microsoft Translator, and SYSTRAN, on a separate in-domain EN-DE test set of 5,136 subtitle segments with independent reference translations (i.e., not post-edits of any system) fetched from the same video streaming platform as the SubEdits corpus. The results (as of May 2020) are summarized in Table \ref{tbl:commercial}.  Unsurprisingly, the proprietary NMT system specialized at translating drama subtitles substantially outperforms other general MT systems. 

\subsection{APE Performance on SubEdits}

\label{sec:ape-perf}

Table \ref{tbl:baseline} reports the performance of vanilla transformer and BERT Enc-Dec APE models and compares it the \textit{do-nothing} NMT baseline (the output produced by the proprietary in-domain NMT system). TF (base) APE improves over the \textit{do-nothing} NMT baseline output ($p<0.05$), particularly on TER scores. However, TF (\textsc{BERT} size) APE shows a smaller improvement on ChrF and TER scores and a drop in BLEU. Even with the SubEdits corpus, large networks such as TF (\textsc{BERT} size) tends to overfit. However, with pre-trained BERT initialization, BERT Enc-Dec APE shows substantial improvement across all metrics. Unlike previous studies that report marginal improvements \cite{chatterjee2018findings,chatterjee2019findings}, our results show that a strong APE model trained on large human post-edits can significantly outperform ($p<0.001$) a strong in-domain NMT system.

\subsection{Human Evaluation}

To validate the improvement in automatic evaluation scores and to estimate the human upper-bound performance on SubEdits, we conducted human evaluation. We hired five German native freelance translators who are also proficient in English and had prior experience with English/German translation.

\begin{figure}[t]
\centering
\includegraphics[width=0.5\textwidth]{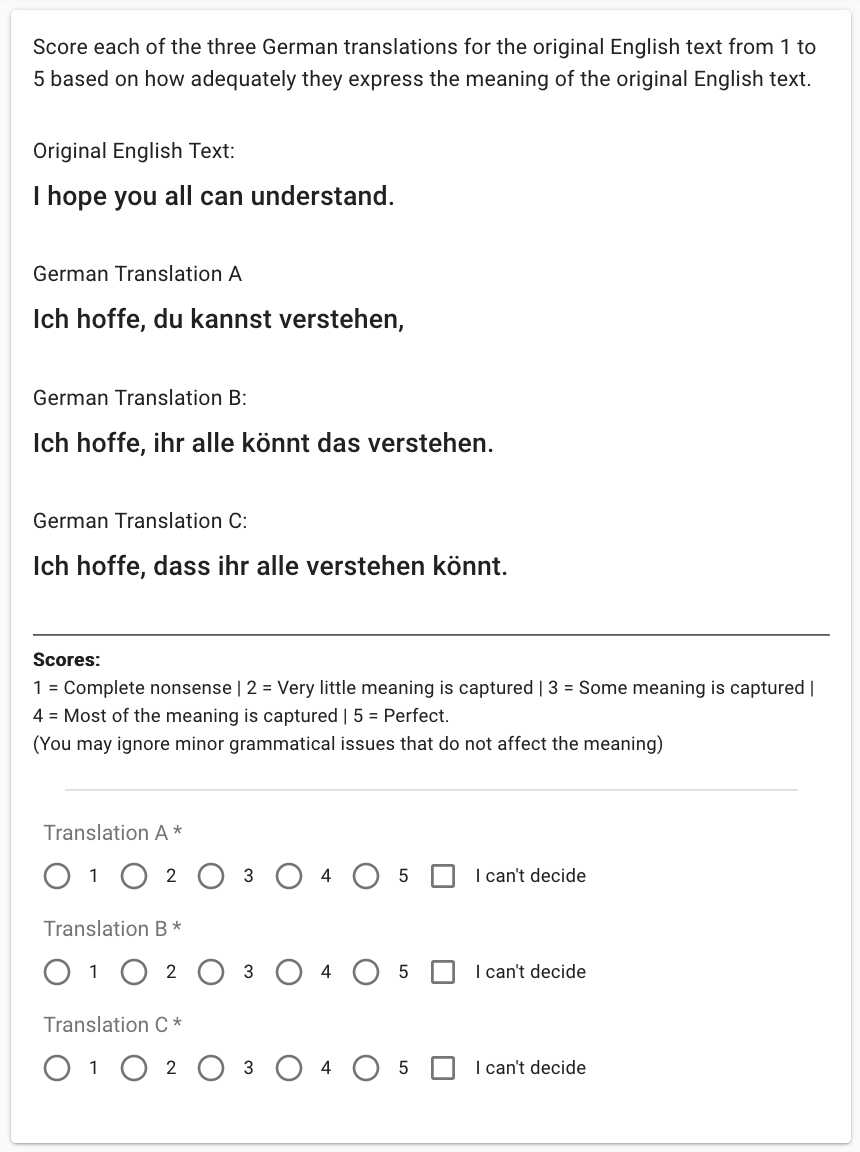}
\caption{Interface used to rate the translations. }
\label{fig:evalscreen}
\end{figure}

Given the original English text, the annotators were asked to rate the adequacy (from 1 to 5) for three German translations: (1) the \textit{do-nothing} baseline output (NMT), (2) BERT Enc-Dec APE output (APE), and (3) the human post-edited text (Human). Figure \ref{fig:evalscreen} shows the interface presented to the annotators for rating the translations. The three translations are presented on the same screen in random order and the annotators are unaware of their origin.

\label{sec:humaneval}

\begin{table}
\centering
\small
\begin{tabular}{|c|cccr|}
\hline
\textbf{Annotator}    &   \textbf{NMT}  &   \textbf{APE} &   \textbf{Human} &   \multicolumn{1}{c|}{\textbf{\ \# Eval.}}\\
\hline
 A           &  3.7 &   4.2 &     4.5 &           593 \ / \ 603 \\
 B           &  3.5 &   4.0 &     4.4 &           594 \ / \ 603 \\
 C           &  3.7 &   4.3 &     4.4 &           603 \ / \ 603 \\
 D           &  2.8 &   3.4 &     3.8 &           587 \ / \ 603 \\
 E           &  3.3 &   3.8 &     4.3 &           602 \ / \ 603 \\ \hline
 A-E         &  3.4 &   3.9 &     4.3 &          2979 \ / 3015 \\
\hline
\end{tabular}
\caption{Average adequacy scores (1-5) rated by annotators (A to E). Overall average is shown in the last row (A-E).  }
\label{tbl:human}
\end{table}

Following recent WMT APE tasks \cite{bojar2017findings,chatterjee2018findings,chatterjee2019findings}, our human evaluation is also based solely on adequacy assessments. Previous studies reported a high correlation of fluency judgments with adequacy \cite{callisonburch2007meta} making the fluency annotations superfluous \cite{przybocki2009nist}. Unlike the recent WMT APE tasks, we did not opt for direct assessments \cite{graham2013continuous} since we wanted to evaluate the degradation or improvement in the quality of the NMT output due to APE and human post-edits on the same English source segments. 

We elicit judgments for all test set instances where the APE model modified the NMT output beyond simple edits on punctuation, HTML tags, spacing, or casing. 2,815 out of the 10,000 instances in our test set contains non-simple edits. A set of 50 instances out of 2,815 was evaluated by all annotators to compute inter-annotator agreement.\footnote{Each annotator scored 603 test instances.}%, $ \frac{2815-50}{5} + 50 = 603$}

After evaluation, we filtered out the instances where the annotator was unable to decide a score for any of the three translations. The average scores by each annotator (A to E) and the overall average scores are shown in Table \ref{tbl:human}. The numerator of the ``\# Eval.'' column indicates the number of evaluations used for the average score computation after filtering out the ``\textit{I can't decide}'' annotations. The results of our human evaluation (Table \ref{tbl:human}) show that all five annotators rate the APE output better than baseline NMT output by at least $+0.5$ on average, reaching an overall score of 3.9.  All the five annotators rated the human post-edited output substantially better than the NMT output and the APE output, which indicates that quality of the post-edits in the SubEdits corpus is high. Human post-edits received an overall average score of 4.3.

Using the repeated set of 46 instances,\footnote{We removed 4 instances out of the 50, where one or more annotators chose the ``I can't decide'' option.} we compute inter-annotator agreement using average pairwise Cohen's Kappa $\kappa$ \cite{cohen1960coefficient} to be 0.27 which is considered to be fair \cite{landis1997measurement} and similar to that observed for adequacy judgments in WMT tasks \cite{callisonburch2007meta}. However, the ranges of scores used by the annotators differ considerably (especially, annotator `D'). Hence, measures such as a weighted Kappa $\kappa_w$ \cite{cohen1968weighted}, which assigns partial credit to smaller disagreements and works better with ordinal data (such as our adequacy judgments), is more suitable.  We compute the average pairwise quadratically weighted Kappa $\kappa_w$ to be $0.50$, and consider their agreement to be moderate.

\begin{figure*}[t]
\centering
\includegraphics[width=\textwidth]{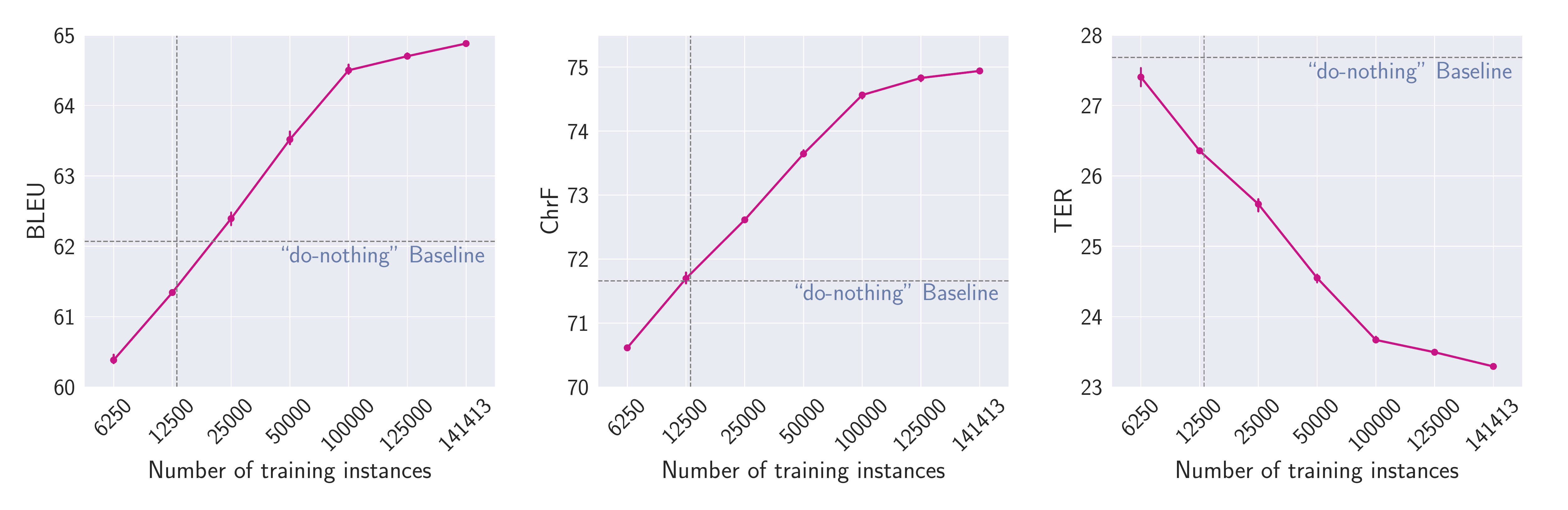}
\caption{Performance of BERT Enc-Dec APE model with varying training data size in terms of BLEU, ChrF, and TER metrics on the SubEdits dev set. The vertical dotted line in each figure shows the data size used for WMT APE EN-DE (NMT) task (13,441 triplets) and the horizontal dotted line shows the NMT Baseline results. }
\label{fig:data-size}
\end{figure*}
\subsection{Can APE substantially improve in-domain NMT with adequate data size?}

To analyze the effect of training data size with respect to APE performance, we train BERT Enc-Dec APE with varying sizes of training data from the SubEdits corpus and evaluated the models on the SubEdits development set. For each training data size, ranging from 6,250 to 125,000, we train three models on three random samples of the respective size from the SubEdits training set. Each point in Figure \ref{fig:data-size} denotes the \textit{mean} score of the three models (the vertical error bars at each point denote the \textit{minimum} and \textit{maximum} scores). The \textit{do-nothing} NMT baseline score is represented by a \textit{horizontal} dotted line. As a reference, we mark the size equivalent to that of WMT'18 APE EN-DE (NMT) training set (13,441 triplets) with the \textit{vertical} dotted line. The rightmost point on each graph represents the score if the full training corpus is used.

Although the sizes of WMT APE dataset and the SubEdits corpus are not directly comparable, we see that size does matter for better APE performance. When the APE model was trained on a subset of SubEdits corpus that is of the same size as the WMT'18 APE training data, it performs worse than the baseline in terms of BLEU score and only marginally improves in ChrF and TER scores (see intersection points of the vertical and horizontal lines in Figure \ref{fig:data-size}).

Interestingly, doubling the amount of training data from 12,500 to 25,000 provides slight BLEU gains above the \textit{do-nothing} baseline and increasing the data size to 50,000 training instances improves the model further by $+$1 BLEU. The curves continue to show an increasing trend. After 100,000 training instances, the data size effect on score improvement slows down. This experiment shows the possibility that previous work on APE for NMT outputs might have reached a plateau simply due to the lack of human post-edited data rather than the limited usefulness of APE models.

\subsection{How much does artificial APE data help?}

Previous work using strong neural APE models \cite{junczysdowmunt2018msuedin,tebbifakhr2018multi} relied predominantly on artificial corpora such as that released by \citet{junczysdowmunt2016loglinear} and the eSCAPE corpora \cite{negri2018escape}. However, artificial post-edits are either generated from monolingual corpora or independent reference translations and they do not directly address the errors made by the MT system that is to be fixed by APE.

We compare the APE model performance when trained on large-scale in-domain and out-of-domain artificial data (in the order of millions of triplets) to training on the human post-edited SubEdits corpus (over 141K human post-edits). As out-of-domain artificial data, we use the eSCAPE EN-DE NMT corpus and filter sentences that have between 0 and 200 characters resulting in 5.3 million triplets. As in-domain artificial data, we generated an artificial APE corpus using the same approach used to create the eSCAPE corpus by decoding the source sentences from the OpenSubtitles2016 parallel corpus \cite{lison206opensubtitles}, which is also from the subtitle domain \footnote{Although both SubEdits and SubEscape are from the subtitle domain, the translations in SubEscape are from \href{www.opensubtitles.org}{www.opensubtitles.org/} whereas the SubEdits post-edits are compiled from Rakuten Viki.} using the same proprietary NMT system we use to create the SubEdits corpus; the corresponding references translations become the artificial post-edits. We use the same filtering criteria and pre-processing methods for SubEdits (Section \ref{subsec:data-filtering} and \ref{sec:prepost}) resulting in 5.6 million artificial triplets. We set aside 10,000 triplets from each artificial corpus and use it as a development set when training solely on the corresponding corpus. We refer to this artificial corpus as SubEscape.\

\begin{table}[t]
\centering
\small
\begin{tabular}{|l|rrr|}
\hline
 & \textbf{BLEU}$\uparrow$ & \textbf{ChrF}$\uparrow$ & \textbf{TER}$\downarrow$  \\
\hline
\textit{do-nothing} NMT &  61.88 & 71.33 & 28.06\\
\hline
\multicolumn{4}{|l|}{\textbf{w/ BERT Enc-Dec APE trained on:}}  \\
\hline
~SubEdits (R)             & 64.53 & 74.71 & 23.72  \\
~eSCAPE (A)               & 52.35 & 65.65 & 31.95  \\
~SubEscape (A)          & 50.51 & 65.89 & 32.78 \\
~~~~$+$ SubEdits 10$\times$ (A+R)     & \textbf{64.59} & \textbf{75.09} & \textbf{23.41} \\

\hline
\end{tabular}
\caption{APE performance on SubEdits test set when trained with real (R) and artificial (A) training corpora.}
\label{tbl:artificial}
\end{table}

We compare the performance of the BERT Enc-Dec APE trained on SubEdits corpus to that when trained on the artificial corpora in Table \ref{tbl:artificial}. We find that training on artificial corpora alone, irrespective of their domain, cannot improve over the \textit{do-nothing} baseline and in fact, degrades the performance substantially. However, when we combine SubEscape with up-sampled (10$\times$) SubEdits corpus, we get a small improvement, particularly in terms of ChrF and TER. 

\subsection{How significant is domain shift for APE?}
\label{sec:wmt-eval}
While NMT performance has been known to be particularly domain-dependant \cite{chu2018survey}, domain shift between NMT and APE training has not been investigated previously. To assess this, we evaluate BERT Enc-Dec APE on the canonical WMT'18 APE EN-DE (NMT) dataset.\footnote{WMT'19 APE task also used the same dataset for benchmarking EN-DE APE systems}. The baseline NMT system and datasets used for the WMT'18 task is from the Information Technology (IT) domain and is notably different from the domain of SubEdits. We experiment with different methods of combining SubEdits (out-domain) with the WMT APE training data (in-domain). For all experiments, we use 1,000 instances held out from the WMT'18 APE training data as the validation set. The results are reported in Table \ref{tbl:wmt}. When trained on SubEdits alone, despite its size, we see that there is a drastic drop in performance compared to training the much smaller WMT APE data alone. When we combine SubEdits with 10$\times$ upsampled WMT APE training data, we observe some improvement, particularly in terms of BLEU ($p<0.05$), over training with WMT APE data alone. These results show that in-domain training data is crucial to training APE models to improve in-domain NMT. 

\begin{table}[t]
\centering
\small
\begin{tabular}{|l|p{0.7cm}p{0.7cm}p{0.7cm}|}
\hline
& \textbf{BLEU}$\uparrow$ & \textbf{ChrF}$\uparrow$ & \textbf{TER}$\downarrow$  \\
\hline
\textit{do-nothing} NMT  & 74.73 & 85.89 & 16.84 \\
\hline
\multicolumn{4}{|l|}{\textbf{w/ BERT Enc-Dec APE trained on:}}  \\
\hline
~WMT'18 APE (I) & 75.08 & 85.81 & 16.88  \\
~SubEdits (O) & 49.05 & 69.48 & 39.30  \\
~~$+$WMT'18 APE (O+I) & 74.93 & 85.90 & 16.92 \\
~~$+$WMT'18 APE 10$\times$ (O+I) & \textbf{75.27} & \textbf{86.08}  & \textbf{16.62} \\
\hline
\end{tabular}
\caption{APE performance with in (I) and out-of-domain (O) training data on WMT APE NMT test set.}
\label{tbl:wmt}
\end{table}

\section{Analysis}

\subsection{Impact of APE with varying NMT quality}

\begin{figure}[t]
\centering
\includegraphics[width=0.5\textwidth]{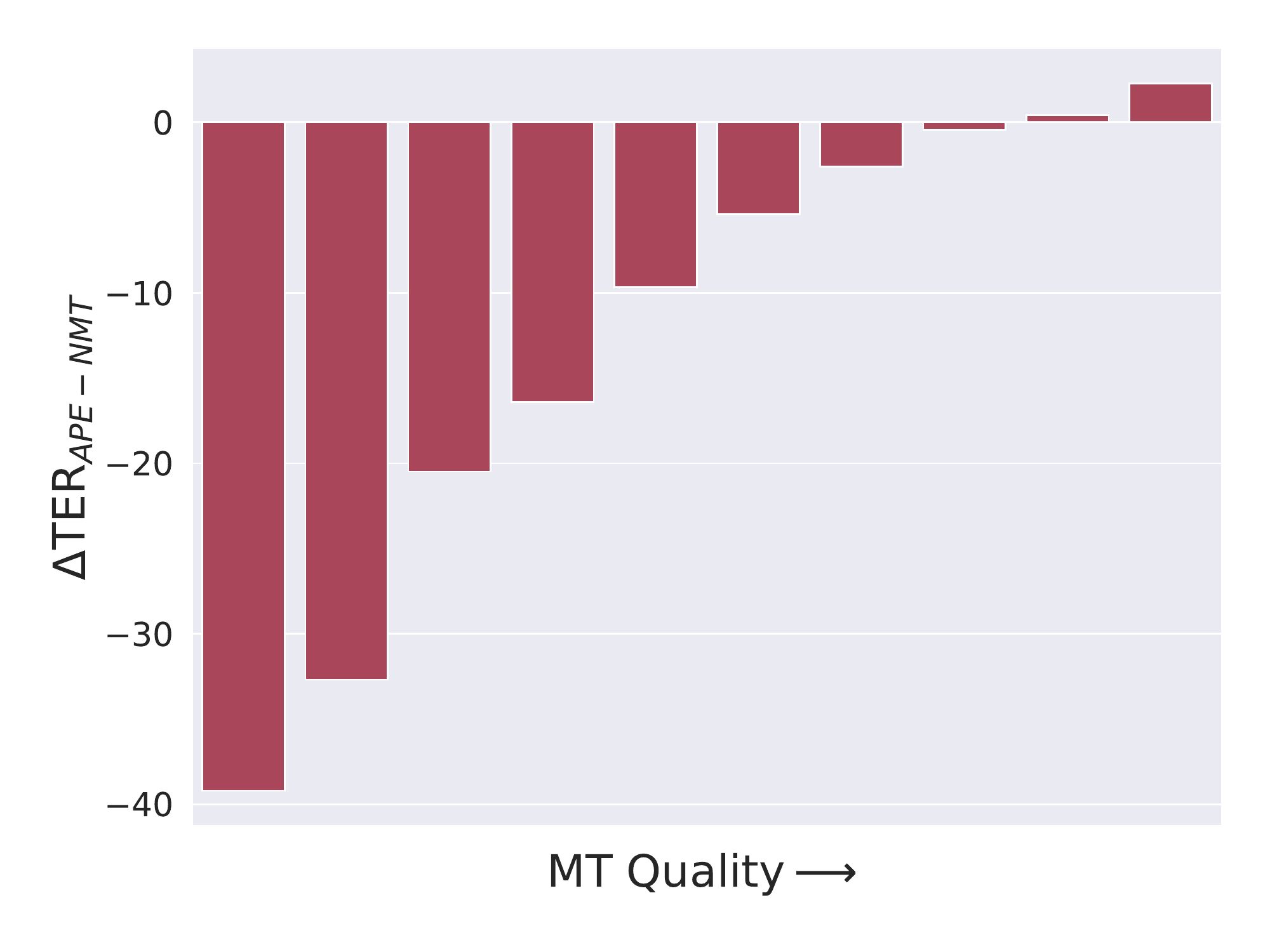}
\caption{Translation quality difference due to APE ($y$-axis) shown by the $\Delta$TER$_{\small \textit{APE}-\textit{NMT}}$ with increasing MT quality ($x$-axis). Negative $\Delta$TER indicates improvement in performance.  }
\label{fig:ape-nmt-quality}
\end{figure}

To study the impact of APE with varying quality of NMT output, we conduct analysis on subsets of our development set with varying translation qualities (Figure \ref{fig:ape-nmt-quality}). We split the SubEdits development set into 10 subsets by aggregating those triplets with the NMT output scoring $>90$ TER (lowest quality), $90-81$ TER,  $\ldots$, $20-11$ TER, and $\leq10$ (highest quality). They are ordered from left to right in the $x$-axis in Figure \ref{fig:ape-nmt-quality} according to increasing MT quality. $y$-axis 
denotes the difference ($\Delta$) between the TER score of APE output and NMT output for each subset. The more negative $\Delta$TER indicates a larger improvement due to APE. We find that on the lower quality subsets, APE improves over NMT substantially. This improvement margin reduces with improving NMT quality and can deteriorate the NMT output when NMT quality is at the highest. This experiment shows that APE contributes to improving overall NMT performance by predominantly fixing poorer quality NMT outputs. The APE model's error will dominate and APE can become counter-productive when NMT output is nearly perfect (i.e., when there are very few or no post-edits done on them as indicated by sentence-level TER scores of $<10$). APE task remains relevant until NMT systems achieve this state, which is still not the case even for strong in-domain NMT systems as indicated by our experiments. 

\subsection{Qualitative Analysis}

We qualitatively analyze the output produced by APE on the SubEdits development set to better understand the improvements and errors made by the APE model. Table \ref{tbl:error-analysis} shows three example outputs produced by the APE model along with the original English text (SRC), the \textit{do-nothing} baseline output (NMT), and the human post-edits (Human).

\begin{table}[t]
\small
\centering
\begin{tabular}{|>{\raggedright\arraybackslash}p{0.8cm}|p{6.1cm}|}
\hline
\multicolumn{2}{|l|}{\textit{Example 1: Incorrect named entities}} \\
\hline
 SRC   & Go to \textbf{Zhongcui Palace}!    \\
 NMT    & Geh zum \textbf{Zhongyuan Palast}! \\
 APE   & Geh zum \textbf{Palast Zhongcui}!  \\
 Human & Geht zum \textbf{Palast Zhongcui}! \\
\hline
\multicolumn{2}{|l|}{\textit{Example 2: Missing phrases}} \\
\hline
 SRC   & Let's go back \textbf{to the resort} and we'll talk it out.  \\
 NMT    & Geh zurück und wir werden reden.                    \\
 APE   & Geh zurück \textbf{zum Resort} und wir werden reden.         \\
 Human & Lass uns zurück \textbf{zum Resort} gehen und darüber reden. \\
\hline 
\multicolumn{2}{|l|}{\textit{Example 3: Requires more context}} \\
\hline
 SRC   & Before coming, City Master negotiated with me.                  \\
 NMT    & Bevor \textbf{er} gekommen ist, hat der Stadtmeister ml cit mir verhandelt. \\
 APE   & Bevor \textbf{wir} kommen, hat die Stadtmeisterin mit mir verhandelt.    \\
 Human & Bevor \textbf{ich} kam, hat die Stadtmeisterin mit mir verhandelt.       \\
\hline

\end{tabular}
\caption{Examples where the APE model proposes changes to the NMT output on the SubEdits test set. The original sentence in English (SRC) and the human post-edit (Human) is also shown.}
\label{tbl:error-analysis}
\end{table}

APE is able to fix incorrect named-entity translations made by the NMT system. 
Example 1 demonstrates an example (``Zhongyuan Palast''$\rightarrow$``Palast Zhongcui'') where the incorrect entity is corrected by the APE model to match the human post-edits. 

NMT often under-translates and misses phrases and the APE models usually can patch these under-translations, e.g. Example 2 where the prepositional phrase ``to the resort''$\rightarrow$``zum Resort'' was missing in the MT outputs and the APE model was able to mend the translation. 

As much as sentence-level APE works well empirically, the lack of context results in erroneous translation by the NMT system where it tries to infer a wrong pronoun and the APE model attempts to assume yet another wrong pronoun, e.g. translating a pronoun-dropped source text in Example 3. Often, the prior or future context from video, audio, or other subtitle instances is necessary to fill these contextual gaps. Sentence-level APE cannot address these issues robustly, which calls for further research on multimodal \cite{deena2017exploring,caglayan2019probing} and document-level \cite{hardmeier2015pronoun,voita2019context} translation and post-editing, especially for subtitles.

\section{Related Work}

Until 2018, APE models were benchmarked on SMT outputs through various WMT APE tasks \cite{bojar2015findings,bojar2016findings,bojar2017findings}. The scale of post-edited data provided by these tasks was in the order of 10,000 to 25,000 triplets. The largest collection of human post-edits, released by \citet{zhechev2012machine}, however, was on SMT and consisted of 30,000 to 410,000 triplets across 12 language pairs. On SMT output, participating systems showed impressive gains even with small training datasets from WMT APE tasks \cite{junczys-dowmunt2017amu,tebbifakhr2018multi}. The results of subsequent APE (NMT) tasks were not as promising with only marginal improvements on English-German and no improvement on English-Russian \cite{chatterjee2019findings}.

Previously, there was no study to assess the necessity of larger human post-edited training data on APE performance on NMT outputs which we address in this paper. APE models were predominantly trained on large-scale artificial data combined with a few thousand human post-edits. \citet{junczysdowmunt2016loglinear} proposed generation of large-scale artificial APE training data via round-trip translation approach inspired from \textit{back-translation} \cite{sennrich2016improving}. They combined artificial training data with real data provided by WMT APE tasks to train their model. Using a similar approach of generating artificial APE data, \citet{freitag2019ape} trained a monolingual re-writing APE model trained on the generated artificial training data alone. Contrary to the round-trip translation approach, large-scale artificial APE data was generated by simply translating source sentences using NMT and SMT systems and using the reference translations as the ``pseudo'' post-edits to create eSCAPE corpus \cite{negri2018escape}. Using the eSCAPE English-Italian APE corpus, \citet{negri2018online} assessed the performance of an online APE model in a simulated environment where the APE model is updated at test time with new user inputs. They found that their online APE models trained on eSCAPE found it difficult to improve specialized in-domain NMT systems. 

Such an analysis by training on artificial corpora may not adequately assess the actual potential of APE since these corpora do not fully cater to the task and can be noisy. The ``synthetic'' post-edits are independent or loosely coupled with the MT outputs, and are often drastically different from the MT output. This makes analyzing APE performance over competitive NMT systems on actual post-edited data an important step in understanding the potential of APE research. Contrary to previous conclusions, our analysis shows that a competitive in-domain NMT system can be markedly improved by a strong neural APE model when trained on sufficient human post-edited training data.

\section{Conclusion}

APE has been an effective option to fix systematic MT errors and improve translations from black-box MT services. However, on NMT outputs, APE has shown hardly any improvement since training has been done on limited human post-edited data. The newly collected SubEdits corpus is the largest corpus of NMT human post-edits collected so far. We reassessed the usefulness of APE on NMT using this corpus.

We showed that with a larger human post-edited corpus, a strong neural APE model can substantially improve a strong in-domain NMT system. While artificial APE corpora help, we showed that the APE model performs better when trained on adequate human post-edited data (SubEdits) compared to large-scale artificial corpora. Finally, our experiments comparing in and out-domain APE show that domain-specificity of training affects APE performance drastically and a combination of in and out-of-domain data with certain upscaling alleviates the domain-shift problem for APE. We find that APE mostly contributes to improving NMT performance by fixing the poorer-quality outputs that still exist with strong in-domain NMT systems. We release the post-editing datasets used in this paper (SubEscape and SubEdits) along with pre/post-processing scipts at PEDRa GitHub repository (\url{https://github.com/shamilcm/pedra})

\section*{Acknowledgements}
We thank the anonymous reviewers for their useful comments. We also thank Rakuten Viki community members who had contributed subtitle post-edits that helped building the SubEdits dataset.

\bibliographystyle{acl_natbib}
\bibliography{emnlp2020}

\end{document}